\relax
\documentclass[letterpaper]{article} % DO NOT CHANGE THIS
\usepackage{aaai21}  % DO NOT CHANGE THIS
\usepackage{times}  % DO NOT CHANGE THIS
\usepackage{helvet} % DO NOT CHANGE THIS
\usepackage{courier}  % DO NOT CHANGE THIS
\usepackage[hyphens]{url}  % DO NOT CHANGE THIS
\usepackage{graphicx} % DO NOT CHANGE THIS
\usepackage{booktabs}
\usepackage{multirow}
\usepackage{makecell}
\usepackage{pdflscape}
\usepackage[table]{xcolor}
\usepackage{natbib}  % DO NOT CHANGE THIS AND DO NOT ADD ANY OPTIONS TO IT
\usepackage{caption} % DO NOT CHANGE THIS AND DO NOT ADD ANY OPTIONS TO IT
\usepackage{diagbox}

\definecolor{aquamarine}{rgb}{0.5, 1.0, 0.83}
\definecolor{applegreen}{rgb}{0.0, 0.5, 0.0}
\definecolor{myred}{rgb}{0.8, 0.0, 0.0}
\definecolor{darkgreen}{rgb}{0.0, 0.4, 0.13}

\usepackage{microtype}

\usepackage{tabularx}
\usepackage{enumitem}
\usepackage{amsmath}
\usepackage{pifont}
\usepackage{amssymb}
\usepackage{color, colortbl}
\usepackage{appendix}
\usepackage{bbold}
\usepackage[T1]{fontenc}
\usepackage{subcaption}
\usepackage{xspace}
\usepackage{comment}

\usepackage[nameinlink,capitalise]{cleveref}

\usepackage[switch]{lineno}

\newcommand{\bluecheck}{\textcolor{blue}{\ding{51}}}%
\newcommand{\redx}{\textcolor{red}{\ding{55}}}%

\newcommand{\rulesep}{\unskip\ \vrule\ }

\definecolor{Gray}{gray}{0.935}

\newcommand\modelname{\textsc{COMeT}}
\newcommand\socialiqa{\textsc{SocialIQa}}
\newcommand\scs{\textsc{StoryCS}}
\newcommand\dynagen{\textsc{COMeT} - DynaGen}
\newcommand\direct{\textsc{COMeT} - Direct}
\newcommand\selftalk{\textsc{Self-Talk}}
\newcommand\atomic{\textsc{Atomic}}

\DeclareMathOperator*{\argmax}{arg\,max}

\title{
Dynamic Neuro-Symbolic Knowledge Graph Construction \\
for Zero-shot Commonsense Question Answering
}

\newcommand\aitwo{$^\diamondsuit$}
\newcommand\uw{$^\spadesuit$}
\newcommand\stanford{$^\heartsuit$}
\newcommand\aspace{\hspace{2.75em}}

\author{
  Antoine Bosselut \aitwo\stanford\aspace
  Ronan Le Bras \aitwo\aspace
  Yejin Choi\aitwo\uw \\
  }
\affiliations {
  \aitwo Allen Institute for Artificial Intelligence \: \: \:
  \stanford Stanford University \\
  \uw Paul G. Allen School of Computer Science \& Engineering, 
 University of Washington \\
 \texttt{\{antoineb,ronanlb,yejinc\}@allenai.org}
}

\date{}

\begin{document}
\maketitle
% \linenumbers
\begin{abstract}

Understanding narratives requires reasoning about implicit world knowledge related to the causes, effects, and states of situations described in text.  
At the core of this challenge is how to access contextually relevant knowledge on demand and reason over it.  

In this paper, we present initial studies toward zero-shot commonsense question answering by formulating the task as inference over dynamically generated commonsense knowledge graphs.
In contrast to previous studies for knowledge integration that rely on \emph{retrieval} of existing knowledge from \emph{static} knowledge graphs,
our study requires commonsense knowledge integration where contextually relevant knowledge is often \emph{not} present in existing knowledge bases. Therefore, we present a novel approach that  \emph{generates} contextually-relevant symbolic knowledge structures on demand using \emph{generative {neural} commonsense knowledge models}. 

Empirical results on two datasets demonstrate the efficacy of our neuro-symbolic approach for dynamically constructing knowledge graphs for reasoning. Our approach achieves significant performance boosts over pretrained language models and vanilla knowledge models, all while providing interpretable reasoning paths for its predictions.

\end{abstract}

\section{Introduction}
\label{sec:intro}

Understanding narratives requires reasoning about 
all the implicit, but trivially inferable, details of a situation based only on what is explicitly stated in text. 
A statement as simple as ``they went to the club'' instantly invokes a bank of commonsense expectations: they had to get dressed, they were going dancing, they likely had drinks, and so forth. 
These reasoning capabilities are missing in most existing neural language understanding models that learn task-specific representations without acquiring rich background knowledge about the social and physical world. 

\begin{figure}[t]
    \centering
    \includegraphics[trim={15.5cm 0cm 11.5cm 0cm},clip,width=\linewidth]{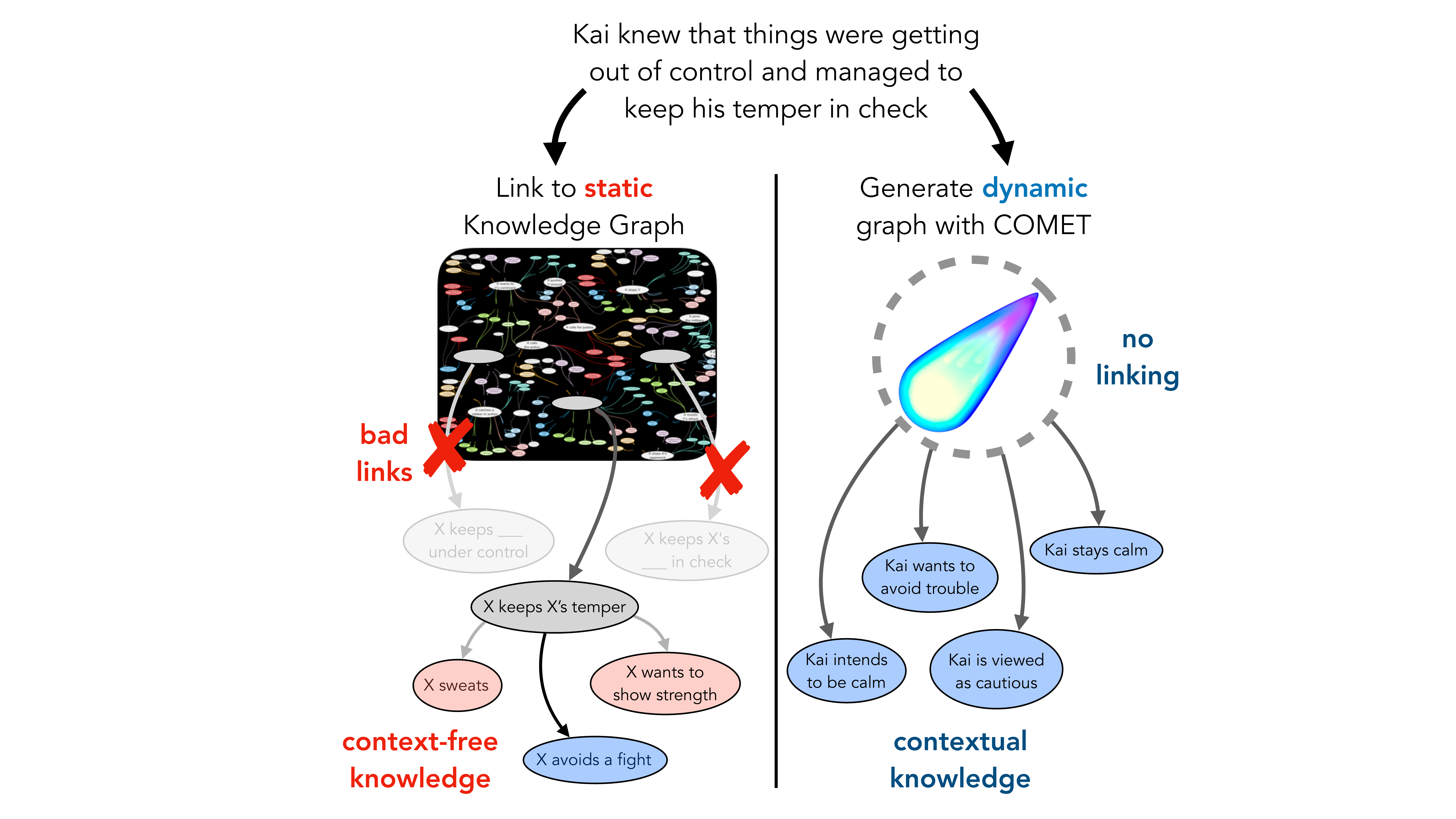}
    \caption{Previous approaches for accessing knowledge link situational contexts to \textbf{static} knowledge graphs. Our work generates knowledge \textbf{dynamically} from neural knowledge models.}
    \label{fig:intro}
\end{figure}

In response, recent work has investigated augmenting deep learning models with retrieval mechanisms over large-scale commonsense knowledge graphs \cite{Mihaylov2018KnowledgeableRE,Bauer2018CommonsenseFG,Paul2019RankingAS}. However, these approaches assume an entity linking step between the written text and knowledge graph. By canonicalizing entities, they discard key context surrounding the input, and often retrieve semantically irrelevant knowledge (e.g., a ``club'' being a blunt weapon is irrelevant to the earlier situation).

In this paper, we propose to \emph{generate} new knowledge that is contextually relevant instead of \emph{retrieving} existing knowledge as is.
\citet{Bosselut2019COMETCT} recently introduced \textit{Commonsense Transformers} (\modelname), a new framework for training neural representations of knowledge graphs. 
This new class of neural \textit{knowledge model} provides a powerful representational tool for connecting commonsense knowledge to downstream task models. 
Because \modelname{} represents knowledge graphs neurally, it can generate commonsense inferences for any entity that can be encoded by the neural model (i.e., described with language). With no need to canonicalize context entities to link to a static knowledge graph, the knowledge model can be queried directly with complex compositional structures, and even full narrative contexts. This flexibility has led them to be used out-of-the-box in a variety of settings requiring contextual knowledge, such as sarcastic comment generation{} \cite{chakrabarty-etal-2020-r}, therapy chatbots{} \cite{Kearns2020AWI}, and story plot generation{} \cite{ammanabrolu2020automated}.

\begin{figure*}[ht]
\begin{subfigure}{.5\textwidth}
  \centering
  % include first image
  \captionsetup{width=.9\linewidth}
  \includegraphics[trim={0cm 4cm 30cm 0cm}, clip,width=\linewidth]{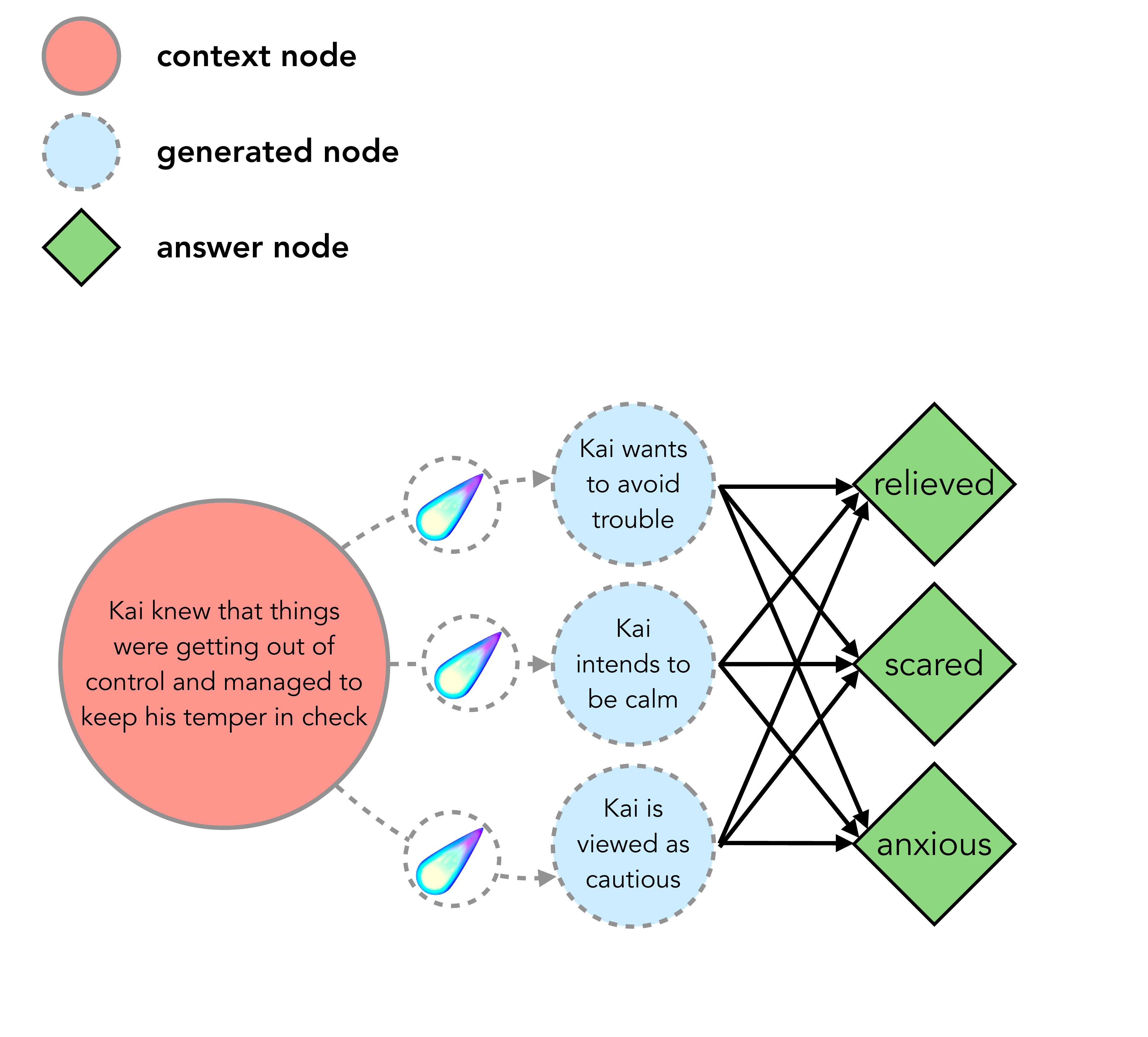}  
  \caption{\modelname{} receives the context $c$ and generates new commonsense inferences to construct a local graph of knowledge about the situation (Section~\ref{sec:construction}).}
  \label{fig:model:construction}
\end{subfigure}
\rulesep
\begin{subfigure}{.5\textwidth}
  \centering
  % include second image
  \captionsetup{width=.9\linewidth}
  \includegraphics[trim={0cm 0cm 25cm 0cm}, clip,width=\linewidth]{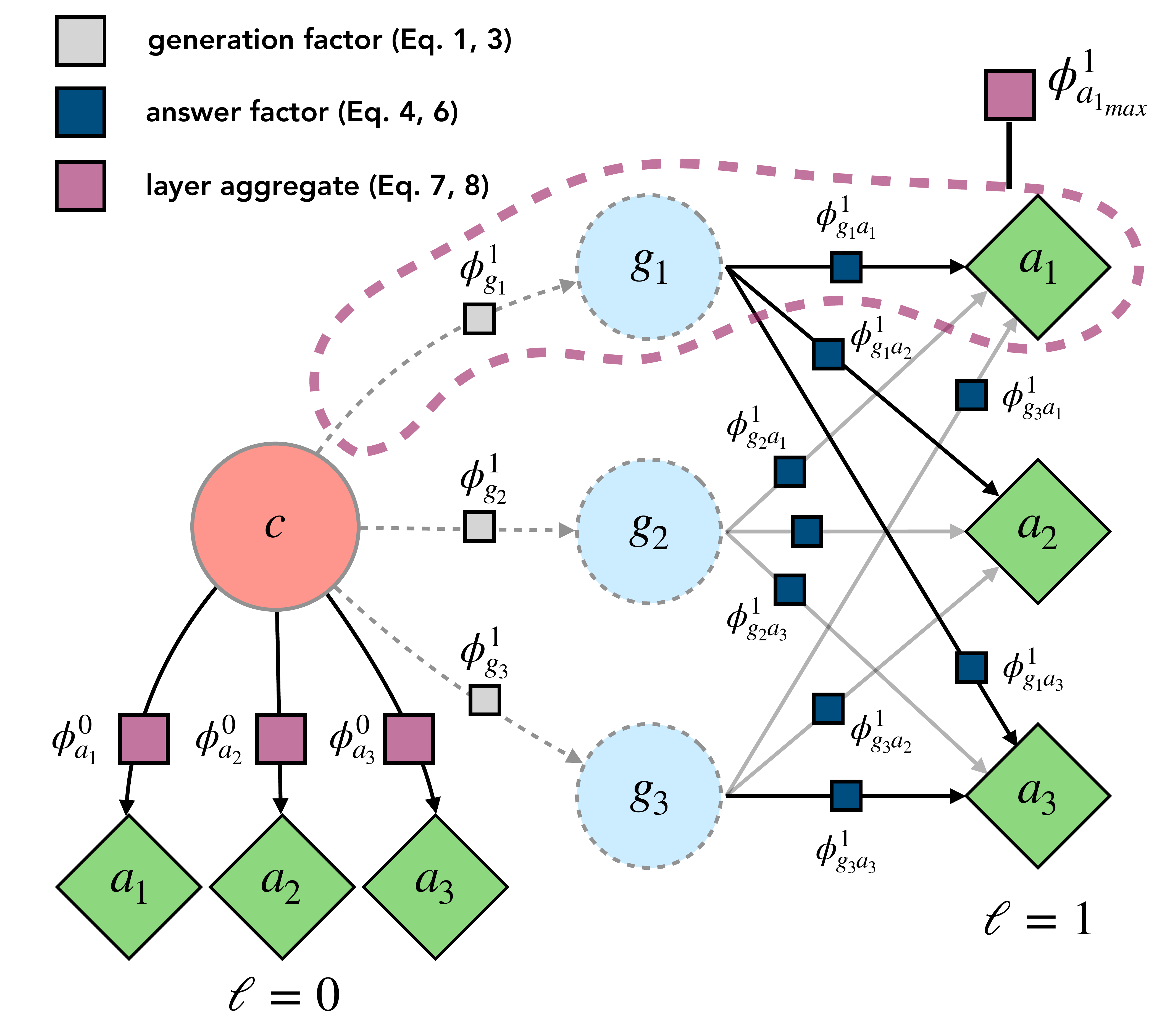}  
  \caption{Our inference algorithms reason over the graph by aggregating commonsense paths to answer questions about the situation (Section~\ref{sec:reasoning}). }
  \label{fig:model:inference}
\end{subfigure}
\caption{Our approach consists of dynamically constructing a local commonsense knowledge graph about a presented situation. This graph can be used to reason about the different questions about the situation.}
\label{fig:model}
\end{figure*}

In this work, we use \modelname{} to construct context-relevant knowledge graphs that can be reasoned over  
for commonsense question answering.  
Given a raw context, \modelname{} generates commonsense inferences that provide world knowledge about the situation depicted in the context. These inferences can be used as additional context to score answer candidates or to generate additional inferences. By generating new inferences and connecting them to the raw context and answers, \modelname{} dynamically constructs a symbolic graph of commonsense knowledge. The raw context is the root node, answer choices are leaf nodes and generated commonsense inferences provide intermediate nodes between them, instantiating different reasoning paths between the context and answers. Using \modelname{} generated scores as factors weighting these paths, we propose new inference algorithms to reason over the generated graph and identify the most likely answers to questions about the situation.

We evaluate our approach in a \textit{zero-shot} setting on the \socialiqa{} \cite{Sap2019SocialIC} benchmark, a question answering dataset for evaluating social commonsense, and the \scs{} benchmark \cite{Rashkin2018psychology}, a story understanding dataset. Empirical results show that our neuro-symbolic approach, \dynagen, outperforms purely neural large-scale pretrained language models \cite{openaigpt,Radford2019LanguageMA} and knowledge models that evaluate QA examples directly without dynamically generating an intermediate symbolic commonsense knowledge graph (i.e., reasoning with \modelname{} with no inference hops). 

\section{Dynamic Knowledge Graph Construction for Question Answering}
\label{sec:construction}
Our approach uses a knowledge model, \modelname~\cite{Bosselut2019COMETCT}, to dynamically construct a context-relevant commonsense knowledge graph about a presented situation. 
\modelname~is trained using transfer learning from large-scale pretrained language models \cite{openaigpt} to knowledge graphs. When trained on the \textsc{Atomic} knowledge graph \cite{sap2018atomic}, it learns to generate social commonsense inferences of situations depicted in text. Importantly, unlike static knowledge graphs (e.g., ConceptNet; \citealp{speer2017conceptnet}), which require canonicalizing input entities to link to the graph, \modelname~represents knowledge neurally, allowing it to generate commonsense for arbitrary input forms. 

In Figure~\ref{fig:intro}, for example, the context ``Kai knew things were getting out of control and managed to keep his temper in check'' is unlikely to be found in any existing knowledge graph. It describes a very specific situation. However, \modelname~can parse this full context and generate commonsense knowledge about Kai's reactions and motivations, such as ``Kai stays calm'' or ``Kai wants to avoid trouble,'' as downstream inferences. We exploit this generalization property of knowledge models to dynamically construct knowledge graphs for presented situations that can be reasoned over to answer commonsense questions about them.

\paragraph{Notation.} Formally, we assume a dataset of examples, each with an associated context $c$ describing a situation, a question $q$ asked about that situation, and a set of $n$ possible answers $\mathcal{A} = \{a^0, ..., a^{n-1}\}$ to that question. Each answer is composed of multiple tokens $Y^{a} = \{y_1, ..., y_{\vert a\vert}\}$. %

\vspace*{2mm}
\noindent
\textbf{Generating Commonsense Inferences.} We generate commonsense inferences for a situational context $c$ by concatenating the context with relation types from the \textsc{Atomic} knowledge graph and using \modelname~to produce candidates $\mathcal{G}$. Each candidate $g \in \mathcal{G}$ is associated with a score $\phi_{g}$ that approximates the model's confidence in the inference:
\begin{equation}
    \phi_{g} = \frac{1}{\vert g \vert}\sum_{t=1}^{\vert g \vert} \log P (x_t | x_{<t} , c, r) \label{eq:cg}
\end{equation}

\noindent where $x_t$ are the tokens of $g$, $\vert g \vert$ is the token length of $g$, $r$ is an arbitrary commonsense relation type for which \modelname~can generate inferences, and:

\begin{equation}
    P (x_t | x_{<t}, c, r) = \mathbb{COMET}(c, r, x_{<t}) \label{eq:comet}
\end{equation}

\noindent where the tokens of $c$ and $r$ are concatenated with the tokens $x_{<t}$ to be input to \modelname. Any generation $g \in \mathcal{G}$ conditioned on $c$ can be seen as a 1-hop commonsense inference of $c$.

Using a Markov assumption, we can generalize this approach by conditioning on generated commonsense inferences to generate $\mathcal{G}^{\ell}$, a set of $\ell$-hop inferences from $c$:

\begin{equation}
 \phi_g^{\ell} = \phi_g^{\ell - 1} + \frac{1}{\vert g^{\ell} \vert}\sum_{t=1}^{\vert g^{\ell} \vert} \log P (x_t | x_{<t} , g^{\ell - 1}, r) \label{eq:cgl}
\end{equation}

\noindent where $\phi_g^{\ell}$ is a generation score for any $g^{\ell} \in \mathcal{G}^{\ell}$, $g^{\ell - 1}$ is an arbitrary inference from $\mathcal{G}^{\ell - 1}$, the set of inferences of the previous hop, and $\phi_g^{\ell - 1}$ is the generation score of that seed inference. Using this approach, we can use \modelname~to construct a graph where commonsense inferences $g$ are nodes. For an arbitrary node $g^{\ell}$, its parent is the node from the previous level $\mathcal{G}^{\ell - 1}$ that \modelname~conditions on to generate $g_{\ell}$. The children of $g_{\ell}$ are nodes generated when \modelname~conditions on $g_{\ell}$ to generate new commonsense inferences. % 
We set $g^0 = c$ because the context is the root node of the graph, and $\phi_g^0 = 0$ because the original context $c$ is deterministic.

\vspace*{2mm}
\noindent
\textbf{Answers as Leaf Nodes.} The final step in constructing the knowledge graph is to connect the answer choices $a \in \mathcal{A}$ to the generated commonsense inferences. We initialize a node in the graph for each answer choice $a$ and connect it as a child node to each commonsense inference in the graph: $g \in \mathcal{G}^{\ell}$ for  $\ell \in [0, L)$ where $L$ is the number of levels in the final graph. In Figure~\ref{fig:model:inference}, we see that the answer choices $\mathcal{A}$~=~\{\textit{relieved, scared, anxious}\} are connected to the root node and each generated commonsense inference in the $L=2$ level graph.

\section{Knowledge Graph Reasoning}
\label{sec:reasoning}
Being designed as a conditional language model, \modelname~can also be used to score candidate commonsense inferences.
%In this work, 
We use this property to score
answer candidates $a \in \mathcal{A}$
conditioned on the generated commonsense inferences $g \in \mathcal{G}$ that are connected to them.
The scores from \modelname~are used to initialize factor nodes between each generated commonsense inference (at all levels of the graph) and each answer choice. Using these scores, and scores between commonsense inferences (Eqs.~\ref{eq:cg}, \ref{eq:cgl}), as a set of factors, our generated knowledge graph implicitly encodes a factor graph that can be reasoned over % 
to evaluate each answer candidate.

\subsection{Computing Answer Scores} 
\modelname~is originally trained to maximize the conditional log-likelihood of the tokens of a target entity $e_2$ from a knowledge graph tuple ($e_1$, $r$, $e_2$). As a result, the knowledge model can measure the log-likelihood of a candidate entity $e_2$ given a source entity $e_1$ and relation $r$. %
For a given example, we treat each answer candidate $a$ as an $e_2$ candidate for \modelname, map the parent nodes of $a$ (e.g., $g$ nodes) to be equivalent to $e_1$, and set the question $q$ as $r$, allowing \modelname~to evaluate each answer candidate according to its implicit knowledge representations. For each answer $a \in \mathcal{A}$, we define a factor based on each token's conditional log-likelihood as computed by \modelname:
\begin{equation}
    \phi_{ga} =  \frac{1}{\vert a \vert}\sum_{s=1}^{\vert a \vert} \log P (y_s | y_{<s} , g, q) \label{eq:ga}
\end{equation}

\noindent where $y_s$ corresponds to the token in $a$ at time step $s$, $y_{<s}$ is all the tokens preceding $y_s$ in $a$, and $\vert a \vert$ is the total number of tokens making up $a$. In this way, for any QA example, we define a set of factor nodes $\phi_{ga}$ connecting the answer candidates $a \in \mathcal{A}$ to the commonsense inferences $g \in \mathcal{G}$ generated by \modelname~about the situational context $c$. 

\vspace*{2mm}
\noindent
\textbf{Overcoming Answer Priors.} Because certain answer candidates have a high probability of occurring for certain questions regardless of the context (e.g., \textit{happy} is a common answer for questions about emotional reactions), we redefine $\phi_{ga}$ (Eq.~\ref{eq:ga}) in terms of the point-wise mutual information between the commonsense path $g$ and answer $a$:

\begin{equation}
    \phi_{ga} \propto \text{PMI}(a,g | q) \nonumber
\end{equation}

\begin{align}
    \phi_{ga} = \frac{1}{\vert a \vert}\sum_{s=1}^{\vert a \vert}\Big( \log& P (y_s | y_{<s} , g, q) \nonumber\\
        &- \log P (y_s | y_{<s}, q)\Big)
\label{eq:loglikelihood_answer_pmi}
\end{align}

\noindent where $\log P (y_s | y_{<s}, q)$ is the log-likelihood of each token in the answer given only the question and previous answer tokens. We describe our approximation of this distribution in Appendix~\ref{app:hyperparameters}.

\subsection{Inference}

Each $\phi_{g}^{\ell}$ scores a unique reasoning path at a particular depth $\ell$ in the graph. The composition $\gamma_{g}\phi_{g}^{\ell} + \gamma_{ga}\phi_{ga}^{\ell}$ can then be seen as scoring a path to a particular answer. To find the most likely answer, we marginalize over all paths to the answers at layer $\ell$:%
%
% \vspace{-5pt}
\begin{equation}
    \phi^{\ell}_{a} = f ( \{\gamma_{g}\phi_{g}^{\ell} + \gamma_{ga}\phi_{ga}^{\ell} : g \in \mathcal{G}^\ell \}) \label{eq:cga_marginal}
\end{equation}
\noindent where % 
$\phi_{g}^{\ell}$ (Eq.~\ref{eq:cgl}) and $\phi_{ga}^{\ell}$ (Eq.~\ref{eq:loglikelihood_answer_pmi}) %
are the \textit{path} and \textit{answer} score, respectively, for generation $g \in \mathcal{G}^{\ell}$. $\gamma_{g}$ and $\gamma_{ga}$ are hyperparameters balancing the contribution of both scores. Because the path and answer scores are log-probabilities, we set $f$ as the LogSumExp, yielding Eq.~\ref{eq:cga_marginal} as a variable elimination over $g \in \mathcal{G}^{\ell}$. 
We also define an extremum estimator over the distribution of generated inferences $\mathcal{G}^\ell$:
\begin{equation}
    \phi^{\ell}_{a_{max}} = \max_{g \in \mathcal{G}^\ell} \gamma_{g}\phi_{g}^{\ell} + \gamma_{ga}\phi_{ga}^{\ell} \label{eq:cga_min}
\end{equation}

\noindent At a high level, $\phi^{\ell}_{a_{max}}$ can be interpreted as approximating the likelihood of answer $a$ given a singular reasoning path: $\{c \rightarrow g^1 \rightarrow \dots \rightarrow g^\ell \rightarrow a\}$, rather than by computing an aggregation of all paths in the graph to the answer (Eq.~\ref{eq:cga_marginal}).

Once the answer scores at different levels in the graph are computed, $\{\phi_{a}^\ell\}^L_0$, the final score for each answer can be evaluated by averaging over the graph levels $\ell \in [0, L)$:% 

\begin{align}
     \log P(a | q, c) \propto \phi_{a} &= \sum_{\ell = 0}^{L} \beta^{\ell} \phi^{\ell}_{a}  \label{eq:ensemble} \\
         {\hat a} &= \argmax_{a \in \mathcal{A}} \phi_{a} \label{eq:answer_ens}
\end{align}

\noindent where $\hat{a}$ is the selected best answer by the approach, $L$ is the number of generation hops made by the \modelname~model (i.e., the number of levels in the graph), $\phi^{\ell}_{a}$ is the score that is propagated from each hop of the constructed knowledge graph, and $\beta^{\ell}$ is hyperparameter scaling the contribution of each hop score. We note that $\phi^{0}_{a}$ is the result from evaluating the answer candidates directly against the original context $c$, and that $\phi^{\ell}_{a}$ is replaced by $\phi^{\ell}_{a_{max}}$ if the extremum estimator (Eq.~\ref{eq:cga_min}) is used instead of variable elimination (Eq.~\ref{eq:cga_marginal}).

\section{Experimental Setup}
\label{sec:train}

We evaluate our approach in a zero-shot experimental setting. It is a well-studied phenomenon that neural methods trained on crowdsourced data often learn to shortcut reasoning to arrive at a correct answer \cite{gururangan-etal-2018-annotation,Lucy2017AreDR}.
We use a zero-shot setting to simulate the model having to reason about situations it has never encountered before, forcing it to construct reasoning graphs from explicit knowledge it can generate (e.g., knowledge learned by \modelname), and precluding it from learning dataset-specific artifacts. As such, we do not use training data to update model parameters. Furthermore, any result presented on the test set does not have hyperparameters tuned on the development set. 

\subsection{Datasets and Processing}
\label{ssec:train:data}
We evaluate our method on two datasets: \socialiqa{} \cite{Sap2019SocialIC} 
and \scs{} \cite{Rashkin2018psychology}. 

\vspace*{2mm}
\noindent
\textbf{\socialiqa.}  The \socialiqa~dataset evaluates a model's ability to understand the social dynamics underlying situations described in short text snippets. Each example in the dataset consists of a context, a question about that context, and three multiple choice answers. An example from the dataset is shown in Figure~\ref{fig:model}. We outline pre-processing steps for the data in Appendix~\ref{app:data}.

\vspace*{2mm}
\noindent
\textbf{\scs.} The \scs{} dataset consists of short 5-sentence stories with annotated motivations and emotional responses whose labels are drawn from classical theories of psychology (e.g., \citealp{plutchik}).
We map the emotion classification task to a QA task by posing an individual question for each emotion label (\textit{disgust, surprise, fear, anger, trust, anticipation, sadness, joy}) that must be predicted for each example. 
We outline this procedure in Appendix~\ref{app:hyperparameters}.

\subsection{Experimental Settings}
\label{ssec:train:settings}
\paragraph{Hyperparameters.} We use most of the same hyperparameters to train the \modelname~model on the \atomic{} knowledge graph as in \citet{Bosselut2019COMETCT}. However, we use GPT2-345M \cite{Radford2019LanguageMA} as the pretrained language model that seeds \modelname~and freeze the position embeddings so we can generalize to longer contexts. We note that the \socialiqa{} dataset is partially derived from \atomic{} knowledge base tuples. However, we do not allow \atomic{} tuples used to seed \socialiqa{} evaluation examples to be used as training examples for \modelname. We provide more details of this splitting in Appendix~\ref{app:data}.
The number of levels in the graph $L$ is set to 2. 
As we operate in the zero-shot setting, we do not tune hyperparameters. For the \socialiqa{} dataset, we set $\gamma_{g} = \gamma_{ga} = 1.0$ and $\beta^{\ell} = 1.0$    $\forall \ell$. For \scs, we do the same except that $\gamma_{g} = 0$.  
Unless stated otherwise, we use argmax decoding to generate inferences from \modelname, and use variable elimination over the graph to select answers.

\vspace*{2mm}
\noindent
\textbf{Prediction.} To predict an answer on the \socialiqa{} dataset, we use Equation~\ref{eq:answer_ens}. Prediction for \scs{} is less straightforward, as the task is originally binary multi-label classification. To make a prediction, we treat $\phi_{a}$ (Eq.~\ref{eq:ensemble}) for each label $j$ independently and select an answer based on whether $\phi_{a, j}$ is above a label-specific threshold, $\kappa^j$.
To avoid violating the zero-shot setting (i.e., tuning thresholds on the development set), we select the threshold using the score at the percentile of the positive label distribution  (e.g., if the \textit{joy} emotion is present for 20\% of examples, we set the threshold at the score of the 20th percentile of the CDF). 
Thresholds are reported in Appendix Table~\ref{tab:kappa_zs} for each label. 
\begin{table*}[t]
\centering
% \small
\resizebox{0.9\textwidth}{!}{
\begin{tabular}{p{0.35\textwidth} | p{0.32\textwidth} | p{0.33\textwidth}}
\toprule
 \textbf{Situation} & \textbf{Most Contributing Paths in Graph} & \textbf{Answers} \\
\toprule
% \multirow{6}{\linewidth}{Kai knew that things were getting out of control and managed to keep his temper in check.} & \multirow{3}{\linewidth}{Kai wants to avoid getting into trouble calm} & a) relieved\\
% && b) scared\\
% && c) anxious\\\cline{2-3}
% & \multirow{3}{\linewidth}{Kai wants to be calm} & a) relieved \\
% && b) scared\\
% && c) anxious\\\cline{2-3}
% \multirow{3}{\linewidth}{How would others feel as a result?} & \multirow{3}{\linewidth}{Kai stays calm} & a) relieved\\
% && b) scared\\
% && c) anxious\\
% \hline \hline
\multirow{4}{\linewidth}{Jesse drove Ash to the airport and dropped them off at the airport with ease.} & \multirow{3}{\linewidth}{\textbf{Jesse wants to go home}} & \textcolor{blue}{\textbf{a) \textit{drained}}} \bluecheck\\
&& %After the airport drop off he 
b) went to the ticket counter\\
&& c) dropped me off at the airport\\\cline{2-3}
& \multirow{3}{\linewidth}{Jesse wanted to be helpful} & a) drained \\
\multirow{2}{\linewidth}{How would Jesse feel afterwards?}&& %After the airport drop off he 
\textcolor{red}{b) \textit{went to the ticket counter}} \redx\\
&& c) dropped me off at the airport\\%\cline{2-3}
% \multirow{3}{\linewidth}{How would Jesse feel afterwards?}& \multirow{3}{\linewidth}{a} & drained\\
% && After the airport drop off he went to the ticket counter\\
% && dropped me off at the airport\\
\hline \hline
\multirow{4}{\linewidth}{After jumping off the roof of his house Quinn had trouble breathing.} & \multirow{3}{\linewidth}{\textbf{Quinn gets hurt}} & \textcolor{blue}{\textbf{a) {\textit{foolish}}}} \bluecheck\\
&& b) patient\\
&& c) light-headed\\\cline{2-3}
& \multirow{3}{\linewidth}{Quinn wants to get medical help} & a) foolish \\
%\multirow{2}{\linewidth}{
How would you describe Quinn?&& b) patient\\
&& \textcolor{red}{c) \textit{light-headed}} \redx\\%\cline{2-3}
% \multirow{3}{\linewidth}{How would you describe Quinn?} & \multirow{3}{\linewidth}{Quinn needs to jump} & a) foolish\\
% && b) patient\\
% && c) \textit{light-headed}\\
\hline \hline
\multirow{4}{\linewidth}{Alex took notice of the children who were singing at the playground.} & \multirow{3}{\linewidth}{\textbf{Alex is happy}} & a) hurt the children\\
&& \textcolor{blue}{\textbf{b) \textit{joy}}} \bluecheck\\
&& c) tell the children to stop\\\cline{2-3}
& \multirow{3}{\linewidth}{Alex wants to go home} & a) hurt the children \\
%\multirow{2}{\linewidth}{
What will happen to Alex?&& b) joy\\
&& \textcolor{red}{c) \textit{tell the children to stop}} \redx\\%\cline{2-3}
% \multirow{3}{\linewidth}{How would others feel as a result?} & \multirow{3}{\linewidth}{Kai stays calm} & a) relieved\\
% && b) scared\\
% && c) anxious\\
\hline \hline
\multirow{4}{\linewidth}{Taylor was close to winning the game. Taylor ran straight for home plate.} & \multirow{3}{\linewidth}{\textbf{Taylor wants to celebrate}} & a) try to get over that they did win\\
&& \textcolor{red}{\textbf{b) \textit{celebrate the win}}} \redx\\
&& c) wanted to score \\\cline{2-3}
& \multirow{3}{\linewidth}{Taylor wants to be home} & a) try to get over that they did win \\
%\multirow{2}{\linewidth}{
\multirow{2}{\linewidth}{What will Taylor want to do next?}&& b) celebrate the win\\
&& \textcolor{blue}{c) \textit{wanted to score}} \bluecheck\\
\bottomrule
\end{tabular}
}
\caption{Example contexts, paths, and answers for the \dynagen{} model on \socialiqa. We \textbf{bold} the predicted answer and its most contributing path. We \textit{italicize} the most likely answer for each path. Incorrect high-scoring answers for a path are highlighted in \textcolor{red}{red} \redx~and correct answers are highlighted in \textcolor{blue}{blue} \bluecheck. We only present a subset of the generated paths. On average, graphs generated using argmax decoding as the graph construction algorithm yield 10.6 nodes and 26.4 edges (Table~\ref{tab:decode}).}
\label{tab:examples}
\end{table*}

% How would you describe Quinn?

\begin{table}[t]
\centering
\begin{tabular}{l  rr}
\toprule
 \textbf{Model} & \textbf{Dev Acc.} & \textbf{Test Acc.} \\
\midrule
Random %\cite{Sap2019SocialIC} 
& 33.3 & 33.3 \\
GPT    & 41.8 & 41.7 \\
GPT2 - 117M & 40.7 & 41.5   \\
GPT2 - 345M & 41.5 & 42.5 \\
GPT2 - 762M & 42.5 & 42.4 \\
\selftalk & 46.2 & 43.9 \\
% \midrule
\direct & 48.7 & 49.0 \\
% \modelname~- DynaGen & \textbf{49.6} &  \textbf{51.9}\\
\dynagen & \textbf{50.1} &  \textbf{52.6}\\
\midrule
BERT-large (sup.) & 66.0 & 63.5\\
RoBERTa-large (sup.) & \textbf{78.1} & \textbf{77.0} \\
\midrule
Human &  86.9 & 84.4 \\
\bottomrule
\end{tabular}
\caption{Accuracy on the development and test sets of \socialiqa. \dynagen{} is our model.}
\label{tab:results}
\end{table}

\section{\socialiqa~Study}
\label{sec:experiments}

% \subsection{\socialiqa} 

\paragraph{Baselines.} As baselines in the \socialiqa~study, we use large-scale pretrained language models: GPT \cite{openaigpt}, GPT2-117M, GPT2-345M, and GPT2-762M \cite{Radford2019LanguageMA}. To adapt these language models optimally to the QA task, question-answer pairs are automatically converted to a templated form, a process we outline in Appendix~\ref{app:hyperparameters}. We also report the results of a model, \direct, that only uses $\phi^0_{a}$ to select answers (i.e., answers are evaluated with respect to the context with no dynamic graph construction). Additionally, we compare against the \selftalk{} model of \citet{Shwartz2020UnsupervisedCQ}, which queries pretrained language models to generate additional details about a presented situation and appends these to the original context. Finally, we report the result of supervised BERT \cite{bert} and RoBERTa \cite{liu2019roberta} models, and random and human baselines from \citet{Sap2019SocialIC}. 
%\ronan{Should we be upfront about the relationship between Atomic, Social-IQa and COMET? Maybe emphasizing that Atomic uses an ``adversarial split'' based on the events / social situations, and that we used the same split to train COMET? Also, another place to alleviate that concern is when describing Table 1.}
% Antoine: In Appendix with preview in dataset description above.

% \paragraph{Ablations} We investigate whether the decoder for generating inferences from \modelname~affects the quality of the generated facts. We present the effect of the following candidate generation algorithms: argmax decoding, beam search with beam size $b=5, 10$ and top-$k$ sampling \cite{fan-etal-2018-hierarchical,holtzman-etal-2018-learning} with $k$ = 5, 10. For each decoding method, we dynamically generate a graph using every candidate produced by the decoder (e.g., argmax decoding produces one candidate, top-10 sampling produces 10 candidates). 
%We also test the effect of the pretrained language model used to seed \modelname. We train additional versions of the knowledge model from GPT, GPT2-117M, and GPT2-762M.

% \footnote{\url{https://leaderboard.allenai.org/socialiqa/submission/bnf45rvumld257e6qutg}}

% \subsection{Results}
\vspace*{2mm}
\noindent
\textbf{Overall Performance.} We report the main results of our \socialiqa~study in Table~\ref{tab:results}. First, our approach achieves an absolute improvement of $\sim$10.2\% over the top performing language model baseline, GPT2-762M, showing the importance of using knowledge models to represent commonsense. Additionally, our approach of dynamically constructing a knowledge graph \textit{on demand} (\dynagen) performs better than using the knowledge model to directly evaluate answers (\direct) by $\sim$3.6\%, highlighting the value in representing more complex reasoning paths. Finally, the improvement over \selftalk{} depicts the benefit of using a structured graphical representation for reasoning compared to one that uses language models to generate additional situational context sentences for conditioning.

%Figure~\ref{fig:approach} shows an example demonstrating how the knowledge graph can help re-score answer options.

We note, however, that the state-of-the-art performance of the supervised BERT and RoBERTa models is significantly higher, meaning there is room for improvement in developing comparable zero-shot approaches to QA. However, one point of interest is that the performance of training BERT with only 5000 training examples (rather than the full 30k) is close (54\%) to the performance of \dynagen, indicating that knowledge models and joint neuro-symbolic solutions are already promising in low-data regimes.

\vspace*{2mm}
\noindent
\textbf{Qualitative Analysis.} In Table~\ref{tab:examples}, we present top reasoning paths from the graphs generated by \dynagen{}. The strength of our approach can be seen in the first example, where the correct answer, \textit{drained}, is more likely to be a feeling associated with wanting ``to go home," a post-condition in the graph generated by \dynagen. In the original context, this condition is implicit. This benefit to leveraging graph reasoning is also seen in the second example, where Quinn's \textit{foolishness} is linked to ``[getting] hurt.'' We note that \direct, RoBERTa-large, and GPT2-345M all answer this question incorrectly, reinforcing the importance of explicit reasoning graphs. 

In the final two examples, we present uninteresting or failure cases. In the first, the model predicts that Alex will experience \textit{joy} after reasoning through the path that he will be ``happy,'' which, while correct, is merely leveraging synonymy. In the final example, we show a case where the model selects an incorrect answer by reasoning through an incorrect path. By recognizing that ``Taylor wants to celebrate'' as a likely post-condition of the context, the model selects an answer that is incorrect. An interesting secondary failure mode in this example is in the second path through the inference ``Taylor wants to be home.'' While this path selects the correct answer, it would not be considered explanatory by humans. In general, we find these cases to be more common in multi-sentence situations. The compositionality of the context makes it more challenging to generate directed inferences, and the factor nodes become less reliable in the graph. We observe that performance on multi-sentence contexts drops by $\sim$5\%.

% \begin{figure}[t]
%     \centering
%     \includegraphics[trim={0cm 0cm 0cm 0cm},clip,width=\linewidth]{figures/hyp_fig_lse.pdf}
%     \caption{\socialiqa~development set performance across different hyperparameter settings of $\beta^\ell$.}
%     \label{fig:hyper}
% \end{figure}
% \label{ssec:atomic:results}

% \paragraph{Tuning effects} To evaluate how our results would have varied if we had tuned hyperparameters, we vary $\beta^{\ell}$ $\forall \ell \in L$ by increments of 0.1 between 0 and 1 and report the results in Figure~\ref{fig:hyper}. Regardless of the values of these hyperparameters, \modelname~- DynaGen was superior to the pretrained language models in every configuration, and better than \modelname ~- Direct (depicted by red line) 80\% of the time. The black line indicates the performance of the purely zero-shot approach, which is one of the better performing configurations, though better results are possible by varying these values. The best performing configurations often have $\beta^{1} \sim 1.5 \beta^{0}$, highlighting the importance of the constructed graph.
% \vspace{5pt}

% \begin{table}[t]
% \centering
% \begin{tabular}{l  rrrrr}
% \toprule
%  \textbf{Model} & G & B5 & B10 & Top5 & Top10 \\
% \midrule
% \textbf{DynaGen$_{max}$}    & 49.6 & 49.1 & 49.1 & 49.0 & 49.4\\
% \textbf{DynaGen}            & 50.1 & 49.5 & 50.0 & 49.0 & 49.3   \\
% \bottomrule
% \end{tabular}
% \caption{Effect of the decoding algorithm for generating commonsense inferences from \modelname. }
% \label{tab:decode}
% \end{table}

\vspace*{2mm}
\noindent
\textbf{Graph Construction Algorithm.} As the quality of the reasoning paths is essential to our approach, we investigate the effect of the inference generation algorithm. We evaluate the following candidate generation algorithms: argmax decoding, beam search with beam size $b=5, 10$ and top-$k$ sampling \cite{fan-etal-2018-hierarchical,holtzman-etal-2018-learning} with $k$ = 5, 10. For each decoding method, we dynamically generate a graph using every candidate produced by the decoder (e.g., argmax decoding produces one candidate, top-10 sampling produces 10 candidates). 

% \begin{table}[t]
% \centering
% \begin{tabular}{l  cc}
% \toprule
% %  \textbf{Decoding Algorithm} & \textbf{DynaGen$_{max}$} & \textbf{DynaGen} \\
% \textbf{Decoding Algorithm}  & $\phi^{\ell}_{a}$ (Eq.~\ref{eq:cga_marginal}) & $\phi^{\ell}_{a_{max}}$ (Eq.~\ref{eq:cga_min}) \\
% \midrule
% Argmax Decoding    & \textbf{50.1}  & \textbf{49.6} \\
% Beam Search - 5    & 49.5           & 49.1          \\
% Beam Search - 10   & 50.0           & 49.1          \\
% Top-$5$ sampling   & 49.0           & 49.0          \\
% Top-$10$ sampling  & 49.3           & 49.4          \\
% \bottomrule
% \end{tabular}
% \caption{Effect of the decoding algorithm for generating commonsense inferences from \modelname.} %All results are on the \socialiqa~development set.}
% \label{tab:decode}
% \end{table}

\begin{table}[t]
\centering
\begin{tabular}{l | rr | cc}
\toprule
%  \textbf{Decoding Algorithm} & \textbf{DynaGen$_{max}$} & \textbf{DynaGen} \\
\textbf{Algorithm}  & \# \textbf{nodes} & \# \textbf{edges} & $\phi^{\ell}_{a}$ & $\phi^{\ell}_{a_{max}}$ \\
\midrule
Argmax Decoding    &  10.6 & 26.4  &  \textbf{50.1} & \textbf{49.6}    \\
Beam Search - 5    & 43.2 & 156.8  &  49.5          & 49.1             \\
Beam Search - 10   & 83.0 & 316.2  &  50.0         & 49.1             \\
Top-$5$ sampling   & 32.0 & 111.9  &  49.0         & 49.0             \\
Top-$10$ sampling  & 59.9 & 223.8  &  49.3          & 49.4             \\
\bottomrule
\end{tabular}
\caption{Development set accuracy for different graph construction techniques. The average number of nodes and edges in the constructed graphs is presented.} %All results are on the \socialiqa~development set.}
\label{tab:decode}
\end{table}

Our results in Table~\ref{tab:decode} show that the performance \dynagen{} is not dependent on the decoding strategy used to dynamically generate the commonsense knowledge graph. %While sampling methods seem to provide a slight improvement over greedy methods for the same number of candidates, this increase does not appear to be significant. 
This result is promising as it shows that the reasoning procedure is robust to variability in the candidate generations (larger graphs will be less precise). However, it also shows that the approach has difficulty using richer dynamically-generated commonsense knowledge representations to answer questions correctly. These results point to the need for future work in developing algorithms that can aggregate larger sets of commonsense inference paths as more expansive knowledge graphs are constructed using more powerful knowledge models.

\section{\scs~Study} 
\label{sec:scs}

\paragraph{Baselines.} As with \socialiqa, we report the results of a random baseline, pretrained language models adapted to the task, and a model that only uses $\phi^{0}_{a}$ to select answers (\direct). As supervised comparison models, we report the performance of several BERT-based models from \citet{gaonkar2020modeling} that are state-of-the-art for the task. %Similar to our setup, these baselines can only access the sentence of the story for which the emotion classification must be made. These models use TF-IDF features, GloVe embeddings \cite{pennington-etal-2014-glove}, LSTMs \cite{hochreiter1997long}, or CNNs \cite{kim-2014-convolutional}, respectively, to encode the sentence. 
% For each baseline, a multi-label linear classifier separately predicts each emotion label from this joint representation. 
 %Finally, we report the performance of a fine-tuned GPT model that has access to the full story.

\begin{table}[t]
    \centering
    \begin{tabular}{l ccc }
    \toprule
        \textbf{Model} & \textbf{P}& \textbf{R}& \textbf{F1} \\ % & \textbf{Acc} \\
    \toprule
        \textbf{Zero-shot CDF-weighted} & \multicolumn{3}{c}{\textbf{No Training Data}}\\ %& \\
        Random & 20.6 & 20.8 & 20.7 \\%& - \\
        GPT         & 34.7 & 36.4 & 35.5  \\
        GPT2 - 117M & 30.8 & 31.8 & 31.3   \\
        GPT2 - 345M & 33.3 & 35.3 & 34.3  \\
        GPT2 - 762M & 35.5 & 37.4 & 36.4 \\
        \direct  & 37.4 & 36.9 & 37.2  \\ %& 83.4\\
        \dynagen & \textbf{38.9} & \textbf{39.3} & \textbf{39.1} \\ % & 83.7\\
        % \midrule
        % \textbf{Zero-shot 50\% positive} & \multicolumn{3}{c}{\textbf{No Training Data}}\\
        % Random & 10.4 & 50.0 & 17.2 \\%& - \\
        % \modelname  &  &  &  \\
        % \modelname~- DynaGen & \textbf{15.0} & \textbf{71.6} & \textbf{24.8} \\ % & 83.7\\

        % \midrule
        
        % \textbf{Few-shot Tuning} & \multicolumn{3}{c}{\textbf{No Training Data}} \\ % & \\
        % %  &&& \\
        % \modelname  & 16.2 & \textbf{60.3} & 25.5 \\ % & 63.5\\
        % %\modelname~- DynaGen & \textbf{18.6} & 52.5 & \textbf{27.5} \\ % & 71.3\\
        % \modelname~- DynaGen & \textbf{19.0} & 56.5 & \textbf{28.4} \\ % & 70.5 \\
        % R:   0.5647932131495228 18860
        % P:   0.1895845940269818 56186
        % F1:  0.28387922074461
        \midrule
        \textbf{Supervised} &&& \\
        % \multicolumn{1}{r}{ TF-IDF} & 20.1 & 24.1 & 21.9 \\% & - \\
        % % \multicolumn{1}{r}{ Glove} & 15.2 & 30.6 & 20.3 \\% & - \\
        % \multicolumn{1}{r}{ LSTM} & 20.3 & 30.4& 24.3 \\% & - \\ 
        % \multicolumn{1}{r}{ CNN} & 21.2 & 23.4 & 22.2 \\% & - \\
        % \multicolumn{1}{r}{ GPT} & \textbf{41.6} & \textbf{50.2} & \textbf{45.5} \\% & 79.3 \\
        BERT & \textbf{65.6} & 56.9 & 61.0 \\
        BERT + LE & 63.1 & 61.7 & 62.4 \\
        BERT + SS & 57.9 & \textbf{76.4} & \textbf{65.9} \\
    \bottomrule
    \end{tabular}
    \caption{Precision, Recall, F1 on the \scs~dataset. Best models in different training settings are \textbf{bolded} }
    \label{tab:scs}
\end{table}

% \begin{table*}[t]
% \centering
% % \small
% \resizebox{\textwidth}{!}{
% \begin{tabular}{p{0.4\textwidth} | p{0.35\textwidth} | p{0.25\textwidth}}
% \toprule
%  \textbf{Situation} & \textbf{Top Reasoning Paths} & \textbf{Answers} \\
% \toprule
% \multirow{4}{\linewidth}{Daniel was excited to get a remote control boat for his birthday. He asked his dad to drive him to the lake to try it out.} & \multirow{3}{\linewidth}{His dad is helpful} & disgusted, angry, sad\\
% && afraid, happy, \textcolor{blue}{\textbf{trusting}} \bluecheck\\
% && excited, surprised\\\cline{2-3}
% & \multirow{3}{\linewidth}{Daniel wants to try something new} & disgusted, angry, sad\\
% %\multirow{2}{\linewidth}{
% \multirow{2}{\linewidth}{How does Daniel feel?}&& afraid, happy, trusting \\
% &&\textcolor{blue}{\textbf{excited}} \bluecheck, surprised\\
% \bottomrule
% \end{tabular}
% }
% \caption{Example contexts, paths, and answers for the \modelname~CGA model on \scs. We show the paths that lead to predicted emotion and \textbf{bold} the emotions predicted from that path. Correct predicted answers are highlighted in \textcolor{blue}{blue} \bluecheck. }
% \label{tab:examples_scs}
% \end{table*}

\begin{table*}[t]
\centering
% \small
\resizebox{\textwidth}{!}{
\begin{tabular}{p{0.41\textwidth} | p{0.30\textwidth} | p{0.29\textwidth}}
\toprule
 \textbf{Situation} & \textbf{Most Contributing Paths in Graph} & \textbf{Answers} \\
\toprule
\multirow{3}{\linewidth}{Daniel was excited to get a remote control boat for his birthday. He asked his dad to drive him to the lake to try it out.} & \multirow{2}{\linewidth}{His dad is helpful} & disgusted, angry, sad, afraid,\\
&&happy, \textcolor{blue}{\textbf{trusting}} \bluecheck, excited, surprised
%&& excited, surprised
\\\cline{2-3}
& \multirow{2}{\linewidth}{Daniel wants to try something new} & disgusted, angry, sad, afraid, \\
%\multirow{2}{\linewidth}{
\textit{How does Daniel feel?}&&happy, trusting , \textcolor{blue}{\textbf{excited}} \bluecheck, surprised\\
\bottomrule
\end{tabular}
}
\caption{Example \scs{} context, high-scoring paths, and answers for our approach. We show which emotions are predicted through which path by \textbf{bolding} them. Correct answers are highlighted in \textcolor{blue}{blue} \bluecheck. As in Table~\ref{tab:examples}, only a subset of paths in the graph generated by \dynagen{} are shown. Generated graphs for \scs{} have on average 8.8 nodes and 19.3 edges.}
\label{tab:examples_scs}
\end{table*}

% Daniel was excited to get a remote control boat for his birthday. He asked his dad to drive him to the lake to try it out.

% trusting
% oReact - personx is helpful
% Thresh: -1.302074670791626
% Path Score: -1.9194376468658447
% xReact Score: -1.4518189430236816
% oReact Score: -2.387056350708008

% excited
% xIntent - personx wants to try something new
% Thresh: 1.038560390472412
% Path Score: 0.8391141891479492
% xReact Score: -0.5841865539550781
% oReact Score: 2.2624149322509766

% Marcus loved to play video games with all of his friends.|One by one they grew into new interests leaving him behind.
% Marcus started feeling lonely, so he started playing online games.

% sad
% xEffect - personx cries
% Thresh: -3.403128147125244
% Path Score: -3.928008556365967
% xReact Score: -5.3791704177856445
% oReact Score: -2.476846694946289

% ===============================================================

% happy
% xIntent - personx wants to be social
% Thresh: 0.36083555221557617
% Path Score: 0.35356616973876953
% xReact Score: 0.48061370849609375
% oReact Score: 0.2265186309814453

% ===============================================================

\vspace*{2mm}
\noindent
\textbf{Overall Performance.} Our results indicate that our zero-shot algorithm, \dynagen{}, significantly outperforms other zero-shot baselines such as language models, including models with twice the number of parameters. %This result is promising as no additional training is used to adapt the \modelname~model to a classification task. Instead, we use the learned neural knowledge model to score the likelihood of tokens corresponding to emotional reactions for characters in the story. 
Importantly, again, we see consistent improvement from dynamically generating a contextual commonsense knowledge graph, rather than directly evaluating the answer choices with \direct. Our full approach yields higher precision, recall, and F1, than the \direct{} baseline. 

% To evaluate the quality of our untuned thresholds from Section~\ref{ssec:train:settings} based on the CDF of the model's scores, we also report the results of our approach if we are allowed to tune the $\kappa$ thresholds on 20\% of the development data (the same amount used for validation in \citealp{Rashkin2018psychology}). We see large gains in recall from this process, causing our performance to exceed even supervised models. 

\vspace*{2mm}
\noindent
\textbf{Qualitative Analysis.} We once again see the benefit of generating a reasoning graph in Table~\ref{tab:examples_scs}.  \dynagen{} is able to select the two correct answers to ``How does Daniel feel?'' leveraging the path through the commonsense inference that ``His Dad is helpful'' to predict that Daniel is \textit{trusting}, and the path through the commonsense inference ``Daniel wants to try something new'' to predict that Daniel is \textit{excited}. However, there is still much room for improvement, as large-scale pretrained language models that are fine-tuned using supervised data perform considerably better on the task.

\begin{table}[t]
    \centering
    \begin{tabular}{l ccc }
    \toprule
        \textbf{Model} & \textbf{P}& \textbf{R}& \textbf{F1} \\ % & \textbf{Acc} \\
    \toprule
        \textbf{Zero-shot} & \multicolumn{3}{c}{\textbf{No Training Data}}\\ %& \\
        CDF-label & \textbf{39.5} & 39.5 & \textbf{39.5}  \\ %& 83.4\\
        % \modelname~- DynaGen & \textbf{19.9} & \textbf{18.8} & \textbf{19.3} \\ % & 83.7\\
        CDF-50 & {25.9} & \textbf{75.0} & {38.5} \\ % & 
        \midrule
        \textbf{Few-shot Tuning} &&& \\
        Tuned from 4 examples & \textbf{31.1} & 54.6 & 39.4 \\
        Tuned from 10 examples & 30.2 & {64.3} & 41.0 \\
        Tuned from 20 examples & 28.6 & \textbf{73.5} & \textbf{41.1} \\
        \midrule
        {20\% development tuning} & 31.2 & {65.1} & {42.2} \\
    \bottomrule
    \end{tabular}
    \caption{Development set Precision, Recall, and F1 of emotion prediction on the \scs~dataset for different strategies for setting prediction thresholds.}
    \label{tab:scs_thresh}
\end{table}

\vspace*{2mm}
\noindent
\textbf{Few-shot Tuning.} To evaluate the quality of our untuned thresholds from Section~\ref{ssec:train:settings} based on the label distribution threshold of the CDF of the model's scores (CDF-label in Table~\ref{tab:scs_thresh}), we also report the results of our approach using different strategies to set thresholds $\kappa$. First, we explore the impact of tuning the $\kappa$ thresholds on varying amounts of the development set data: 4 examples, 10 examples, 20 examples, and 20\% of the development data (the same amount used for validation in \citealp{Rashkin2018psychology}). In each of these settings, we run a study with 5 different randomly selected sets of examples, and report the average performance. We also report the performance of using the 50$^{{th}}$ percentile score of the CDF as the threshold (CDF-50). In Table~\ref{tab:scs_thresh}, we observe large recall gains from these tuning strategies at the expense of precision. However, tuning using merely 10 examples achieves higher F1 than the default strategy, showing the potential of relaxing to a few-shot setting when limited examples are available.
\section{Related Work}
\label{sec:related}

\paragraph{Question Answering with Knowledge Graphs}

Previous work has explored integrating reasoning over static knowledge graphs for question answering and story understanding.  In general, these approaches extract knowledge tuples from the static KG by linking canonicalized entities to nodes
and performing multi-hop inference along relation paths to form full tuples that can be encoded by a downstream neural architecture \cite{Mihaylov2018KnowledgeableRE,Bauer2018CommonsenseFG, Weissenborn2018DynamicIO, Lin2019KagNetKG,Paul2019RankingAS}.
Similar to our approach of discovering reasoning chains between contexts and answers, \citet{Paul2019RankingAS} extract reasoning paths in ConceptNet between normalized entities from the context answer candidates, but can only discover paths through nodes in the static knowledge graph. Finally, there exists works that also dynamically construct latent knowledge graphs \cite{Das2018BuildingDK,Bosselut17}, but these works presuppose a fixed set of entities that can be KG nodes and then approximate graph edges with neural transformations. In contrast, our algorithm can generate arbitrary nodes, thereby constructing a unique graphical structure for any example.% 

\vspace*{2mm}
\noindent
\textbf{Multi-hop Reading Comprehension} Similar in spirit to reasoning over knowledge graphs for question answering is work in multi-hop reading comprehension. Many datasets for learning to aggregate facts without graph structure have been released in recent years \cite{Weston2015TowardsAQ,welbl-etal-2018-constructing,Yang2018HotpotQAAD, talmor-berant-2018-web}. Approaches designed for these resources generally use large-scale neural networks to attend over supporting facts across text \cite{Zhong2019CoarsegrainFC,Dhingra2018NeuralMF}. Most similar to our work are approaches that construct real-time entity mention graphs as neural reasoning paths \cite{Cao2018QuestionAB,Jiang2019ExplorePA,Jiang2019SelfAssemblingMN,Fan2019UsingLK}.
Our approach differs from these models in that we \textit{generate} relevant supporting information rather than mining it from accompanying documents and conduct our study in a zero-shot setting with no additional training.

\vspace*{2mm}
\noindent
\textbf{Automatic Commonsense KG Construction}
Multi-hop reasoning over commonsense inferences requires construction of knowledge resources and recent approaches have investigated how to mine commonsense knowledge from deep learning models. \citet{sap2018atomic} investigated whether LSTM models could generate new tuples for the \textsc{Atomic} knowledge graph. Similarly, \citet{li2016commonsense} and \citet{saito2018commonsense} explored whether neural models could be used to validate proposed knowledge rather than generating it. \citet{jastrzebski-etal-2018-commonsense} built on these approaches for evaluating novel commonsense knowledge mined from Wikipedia. More recent work mapped commonsense tuples to natural language with templates and used pretrained language models to validate them \cite{davison-etal-2019-commonsense, petroni-etal-2019-language}. Concurrently, other research has explored using pretrained language models and adapting them as generative knowledge graph constructors \cite{Bosselut2019COMETCT, Malaviya2019ExploitingSA}. In contrast to these works that augment static knowledge graphs, our approach focuses on constructing knowledge graphs \textit{on demand} to provide context-dependent commonsense for downstream inference.

\section{Conclusion}

Our neuro-symbolic approach uses neural representations of large-scale commonsense knowledge graphs (\modelname) to generate contextual knowledge graphs \textit{on demand} for zero-shot question answering. Our approach dynamically constructs a knowledge graph of commonsense inferences related to a presented context and uses it to evaluate answer options for a posed question. A novel inference algorithm reasons over the constructed graph to select the most likely answer to a question. Our approach shows promising results at answering questions without training on the end task on two datasets, \socialiqa{} and \scs{}, outperforming zero-shot pretrained language models. Finally, our analysis indicates that dynamically generating a contextualized commonsense knowledge graph for inference performs better than using vanilla knowledge models (\direct) to directly answer questions. 

\section*{Acknowledgments}
We thank Maarten Sap, Hannah Rashkin, Vered Shwartz, and Chandra Bhagavatula for helpful feedback. This research was supported in part by NSF
(IIS-1524371,  IIS-1714566),  DARPA  under  the
CwC program through the ARO (W911NF-15-1-
0543), DARPA under the MCS program through
NIWC Pacific (N66001-19-2-4031), JD.com, and the Allen Institute for AI (AI2).

\bibliography{aaai21}

% \maketitle
\appendix

\section{Datasets and Preprocessing}
\label{app:data}

\subsection{Datasets}
\paragraph{Dataset Statistics} We report statistics of the \socialiqa~and \scs~datasets in Table~\ref{tab:datasets} below:

\begin{table}[h]
    \centering
    \begin{tabular}{l  r r}
        \toprule
        \textbf{Dataset} & \textbf{\# dev} & \textbf{\# test} \\
        \toprule
        \socialiqa{} &  1952 & 2217\\
        \scs & 202360 & 182160 \\ 
        \bottomrule
    \end{tabular}
    \caption{Dataset statistics for the \socialiqa~and \scs~datasets.}
    \label{tab:datasets}
\end{table}

\noindent We use the original dataset splits proposed by the authors. We filter 2 and 7 examples from the \socialiqa{} development and test sets, respectively, that are spam.

\paragraph{\atomic{} and \socialiqa} During its construction, \socialiqa{} was seeded with \atomic{} triples during its curation. We address whether this could be a potential source of bias that benefits the approaches based on \modelname. In our analysis, we find there is \textbf{minimal opportunity for data leakage between these resources.} 

First, the \atomic{} knowledge graph was designed with the idea in mind that it could be trained on using neural models to transfer learn knowledge from language.  As a result, to evaluate transfer in this setting, the  knowledge graph is split into a training, development, and test knowledge graph. These splits were made adversarially, meaning no head entities in the training knowledge graph are found in the evaluation knowledge graphs. The \socialiqa{} evaluation sets maintain this split in their design (\socialiqa{} training set seeded by \atomic{} training KG, etc.). As a result, no example in the \socialiqa{} evaluation sets is derived from a tuple in the \atomic{} training knowledge graph. Our \modelname{} implementation is only trained on the training portion of the \atomic{} knowledge graph, meaning our method does not learn from \textit{any} examples used to design the \socialiqa{} evaluation sets. In our work, we do not use any examples from the 
\socialiqa{} training set. 

Second, the \socialiqa{} authors state that crowdworkers heavily re-edited the \atomic{} triples to generate contexts, questions, and answers for each \socialiqa{} example. In any case, to evaluate whether unintentional overlap could still remain, we ran an analysis to recover close \atomic{} training tuples for each example in the \socialiqa{} development set. We removed stopwords from events, stemmed their tokens, and checked whether they could be recovered in the stemmed tokens of \socialiqa{} contexts. Among recovered \atomic{} events, we checked whether any of their associated tail entities were present in the answer choices of the \socialiqa{} example. 

Using this matching scheme, we found an overlap for only $\sim$1.7\% of examples (34/1954 examples in the development set, largely from the fact that the stemming causes compression that makes events appear to be a subset of a \socialiqa{} example). Furthermore, the \dynagen{} and \direct{} models would still perform better than all baselines on the \socialiqa{} development set with these examples removed. Finally, we note that this level of leakage falls far short of the 30\% leakage identified in commonly-used QA datasets \citep{Lewis2020QuestionAA}.

\section{Additional Experimental Settings}
\label{app:hyperparameters}

\paragraph{Approximating the Marginal Distribution} We approximate the marginal distribution for the PMI calculation in Equation~\ref{eq:loglikelihood_answer_pmi} using Equation~\ref{eq:comet}, but set $c = $ ``\texttt{PersonX}''. Every training example in the \textsc{Atomic} knowledge graph on which \modelname~is trained begins with this token, so using it as the only token in the context essentially provides an output distribution that is only conditioned on the question $q$.

\paragraph{Generation Processing} To ground the conditional distribution on which \modelname~and GPT2 (for baselines) were trained we process the data and generations in the following ways:

\begin{itemize}
    \item  For language model baselines (i.e., the class of GPT2 models), we adapt the QA task as natural language statements to be evaluated by the language models. Question-answer pairs are automatically converted to a templated form. For example, a question such as "How does Alice feel after?" will be replaced by the template ``Alice feels" and prepended to the answer. The resulting snippet is then concatenated to the context, and the language models score the answer words conditioned on the context and template. We record the perplexity of each statement and select the lower perplexity score as the answer. Table~\ref{tab:templates} provides the template for each question variety.
    \item When converting generated inferences to contexts for answer scoring (Eq.~\ref{eq:ga}), we add a prefix that is specific to the inference type to the generated tokens (e.g., happy $\Rightarrow$ Person is happy).
    \item We append the following prefixes to \modelname-generated inferences when using them in Equation~\ref{eq:ga} to compute factor nodes between them and answer nodes:
\end{itemize} 

\begin{table}[h]
    \centering
    \small
    \begin{tabular}{l|l}
    \toprule
    \textbf{Relation} & \textbf{Prefix}\\
    \toprule
    \texttt{xWant} &  PersonX wants \\
    \texttt{xReact} &  PersonX is \\
    \texttt{xNeed} &  PersonX needs \\
    \texttt{xIntent} &  PersonX wants \\
    \texttt{xAttr} &  PersonX is \\
    \texttt{xEffect} &  PersonX \\
    \texttt{oReact} &  PersonX is \\
    \texttt{oEffect} &  PersonX \\
    \texttt{oWant} &  PersonX wants \\
    \bottomrule
    \end{tabular}
    \caption{Prefixes appended to \modelname-produced commonsense inferences for the evaluation step (Eq.~\ref{eq:ga})}
    \label{tab:prefix}
\end{table}

\begin{table*}[t]
    \centering
    \small
    \begin{tabular}{l|l}
    \toprule
        \textbf{Question} & \textbf{Template}  \\
        \toprule
         What will happen to Others? & The effect on others will be \_\_\_\\
        How would Others feel as a result? & Others feel \_\_\_\\
        What will Others want to do next? & After, others will want to \_\_\_\\
        How would you describe CHARACTER? & CHARACTER is \_\_\_\\
        What will happen to CHARACTER? & The effect on CHARACTER will be \_\_\_\\
        What does CHARACTER need to do before this? & Before, CHARACTER needs to \_\_\_\\
        Why did CHARACTER do this? & CHARACTER did this because \_\_\_ \\
        How would CHARACTER feel afterwards? & CHARACTER feels \_\_\_ \\
       What will CHARACTER want to do next? & After, CHARACTER will want to \_\_\_ \\
        \bottomrule
    \end{tabular}
    \caption{Templates used to convert question answering pairs from \socialiqa~to a format that can be evaluated by the baseline pretrained language models: GPT, GPT2-117M, GPT2-345M, and GPT2-762M.}
    \label{tab:templates}
\end{table*}

\begin{itemize}
    \item For the \scs~dataset, when scoring the answer text, we use formulations of the words that make up the classification label (e.g., \textit{disgust, surprise, fear, anger, trust, anticipation, sadness, joy} $\Rightarrow$ \textit{disgusted, surprised, afraid, angry, trusting, excited, sad, happy}). As question representations $q$ to give to \modelname, we use the relations from \textsc{Atomic} \cite{sap2018atomic} that correspond to reactions to events: \texttt{xReact} and \texttt{oReact}. 
We compute $\phi_{ga}$ (Eq.~\ref{eq:ga},~\ref{eq:loglikelihood_answer_pmi}) for each $q$ and average them.

    \item For our main models and ablations, names that appear in contexts and answers are anonymized. 
\end{itemize}

\paragraph{Rules for pruning generation sets} We use the following rules to prune the set of commonsense inferences generated by \modelname~as it constructs a graph of commonsense knowledge:

\begin{enumerate}
    \item Any generation that is ``none" is pruned
    \item Any generation that is identical to a previous generation from the same inputs, but has added punctuation is pruned (e.g., to go to the mall vs. to go to the mall\textbf{.})
    \item Any generation that has the phrase ``PersonY" for the following relations is removed: \texttt{oEffect, oReact, oWant}. These generations are untrustworthy as they are often impossible to resolve with an actual person in the context.
    \item Any generation for the following relations that does not have a token that is a verb is removed: \texttt{xEffect, oEffect}
    \item In multiple candidate settings (i.e., beam search, top-$k$ sampling), if one of the candidates is ``none,'' we prune all candidates with less likely scores.
    \item For the \scs~dataset, we only generate inferences along the following \textsc{Atomic} relations: \texttt{xReact}, \texttt{oReact}, \texttt{xEffect}, \texttt{oEffect}, \texttt{xIntent}. The logic for pruning \texttt{xWant}, \texttt{oWant}, \texttt{xNeed}, \texttt{xAttr} inferences is that emotional reactions for these dimensions could be irrelevant to the context. For example, the emotional reaction to \textit{getting into a car accident} is different from \textit{needing to own a car} to do this. Emotional reactions to the kept relations are more likely to be relevant to the original context.
\end{enumerate}

\newpage
\paragraph{Prediction thresholds} We set the following $\kappa$ thresholds to make positive predictions on the \textsc{StoryCommonsense} dataset.

\begin{table}[h]
    \centering
    \small
    \begin{tabular}{l r r }
    \toprule
        \multirow{2}{*}{\textbf{Dimension}} & \multicolumn{2}{c}{\textbf{CDF - label}} %& %\multicolumn{2}{c}{\textbf{Strategy \#2}} 
        \\
        
        &\textbf{Direct $\kappa$} & \textbf{DynaGen $\kappa$}  \\
        \toprule
        disgust      & 5.878  & 6.272 \\
        surprise     & 4.790  & 5.452 \\
        fear         & 6.504  & 6.640 \\
        anger        & 3.773  & 4.093 \\
        trust        & 8.064  & 8.126 \\
        anticipation & 3.765  & 4.008 \\
        sadness      & 3.473  & 3.548 \\
        joy          & 1.907  & 1.913 \\
        \bottomrule
    \end{tabular}
    \caption{Percentile thresholds  $\kappa$  for predicting an emotion for the \direct{} and \dynagen{} models}
    \label{tab:kappa_zs}
\end{table}

\begin{table*}[t]
\centering
\small
\begin{tabular}{p{2.0cm}p{4.7cm}p{5.5cm}}
\toprule

 \textbf{Relation} & \textbf{Description} & \textbf{Example Completion:}     \\
 \midrule
 & & \textbf{Event: }Person X puts Person X's trust in Person Y \\
\midrule 
\texttt{oEffect}    & The effect the event has on others besides Person X &  is considered trustworthy \newline is believed \newline gains Person X's loyalty  \\
\midrule
\texttt{oReact}     & The reaction of others besides Person X to the event  &  trusted\newline honored\newline trustworthy \\
\midrule
\texttt{oWant}      &  What others besides Person X may want to do after the event & work with Person X\newline partner with Person X\newline to help Person X \\
\midrule
\texttt{xAttr}      &  How Person X might be described given their part in the event &  faithful\newline hopeful\newline trusting\\
\midrule
\texttt{xEffect}    &  The effect that the event would have on Person X & gets relieved\newline stays faithful\newline Is betrayed\\
\midrule
\texttt{xIntent}    & The reason why X would cause the event & to be trusting\newline his or her help/guidance/advice\newline to be friends \\
\midrule
\texttt{xNeed}      &  What Person X might need to do before the event & to be friends with Person Y\newline to have heard a lot of good things about Person Y\newline to get to know Person Y\\
\midrule
\texttt{xReact}     &  The reaction that Person X would have to the event &  trusting\newline safe, not alone \newline understood\\
\midrule
\texttt{xWant}      &  What Person X may want to do after the event& to rely on Person Y\newline to go into business with Person Y\newline to make sure that their heart feeling is right \\
\midrule
\end{tabular}
\caption{Definitions of the relations in \textsc{Atomic}. Events in \textsc{Atomic} center around the personal situations of a central figure, Person X, with potentially more participants.}
\label{tab:app:dimensions}
\end{table*}
\end{document}